\documentclass[10pt,twocolumn,letterpaper]{article}

\usepackage{wacv}              

\usepackage{graphicx}
\usepackage{amsmath}
\usepackage{amssymb}
\usepackage{booktabs}

\usepackage{times}
\usepackage{epsfig}
\usepackage{multirow}
\usepackage{ulem}
\usepackage{capt-of}
\usepackage{xcolor}
\usepackage{colortbl}
\usepackage[accsupp]{axessibility}

\usepackage[pagebackref,breaklinks,colorlinks]{hyperref}

\usepackage[capitalize]{cleveref}
\crefname{section}{Sec.}{Secs.}
\Crefname{section}{Section}{Sections}
\Crefname{table}{Table}{Tables}
\crefname{table}{Tab.}{Tabs.}

\def\wacvPaperID{507} 
\def\confName{WACV}
\def\confYear{2025}

\begin{document}

\title{Revisiting Disparity from Dual-Pixel Images:\\
Physics-Informed Lightweight Depth Estimation}
\author{Teppei Kurita \qquad Yuhi Kondo \qquad Legong Sun \qquad Takayuki Sasaki \qquad Sho Nitta \\
\qquad Yasuhiro Hashimoto \qquad Yoshinori Muramatsu \qquad Yusuke Moriuchi\\
Sony Semiconductor Solutions Corporation\\
\tt \small\{Teppei.Kurita, Yuhi.Kondo, Legong.Sun, Takayuki.Sasaki, Sho.Nitta, \\
\tt \small Yasuhiro.Hashimoto, Yoshinori.Muramatsu, Yusuke.Moriuchi\}@sony.com
\\
\small \url{https://github.com/sony/dual-pixel-disparity}
}

\twocolumn[{
	\renewcommand\twocolumn[1][]{#1}
	\maketitle
	\begin{center}
		\vspace{-0.7cm}
        \begin{tabular}[b]{c}
          \includegraphics[height=1.78in]{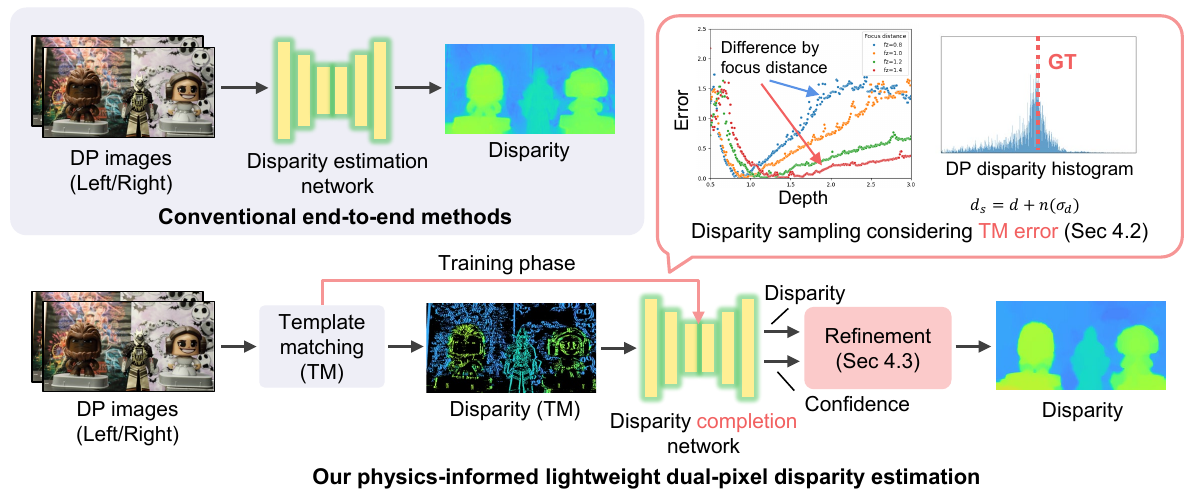}\\
          \small (a) Comparison of proposed and conventional end-to-end methods
        \end{tabular}
		\hspace{-0.45cm}
        \begin{tabular}[b]{c}
          \includegraphics[height=1.78in]{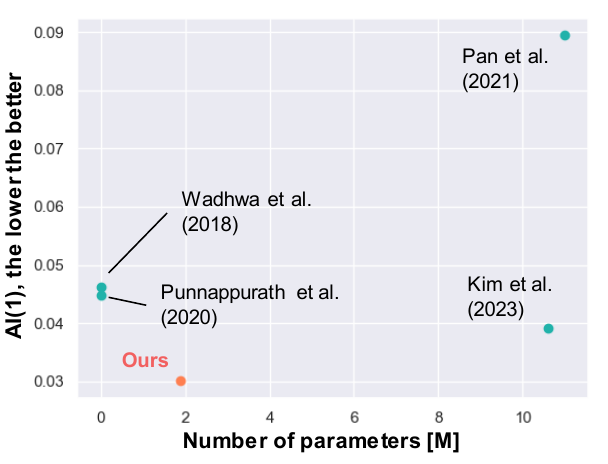}\\
          \small (b) Disparity estimation performance
        \end{tabular}
        \vspace{-0.3cm}
        \captionof{figure}{
		\textbf{Overview of our method.}
(a) Conventional end-to-end disparity or depth estimation from dual-pixel (DP) images is redundant and limited in performance because it does not exploit disparity constraints explicitly. The proposed method efficiently estimates disparity using a completion network trained to consider errors during template matching and a framework that refines the expansion of disparity regions that occur in principle.
(b) shows the number of parameters on the horizontal axis and accuracy on the vertical axis; the closer to the lower left, the better the performance balance. The proposed method is lightweight and achieves high performance.
		}
	\label{fig:teaser}
	\end{center}
}]

\begin{abstract}
\vspace{-0.3cm}
In this study, we propose a high-performance disparity (depth) estimation method using dual-pixel (DP) images with few parameters.
Conventional end-to-end deep-learning methods have many parameters but do not fully exploit disparity constraints, which limits their performance.
Therefore, we propose a lightweight disparity estimation method based on a completion-based network that explicitly constrains disparity and learns the physical and systemic disparity properties of DP.
By modeling the DP-specific disparity error parametrically and using it for sampling during training, the network acquires the unique properties of DP and enhances robustness.
This learning also allows us to use a common RGB-D dataset for training without a DP dataset, which is labor-intensive to acquire.
Furthermore, we propose a non-learning-based refinement framework that efficiently handles inherent disparity expansion errors by appropriately refining the confidence map of the network output.
As a result, the proposed method achieved state-of-the-art results while reducing the overall system size to $1/5$ of that of the conventional method, even without using the DP dataset for training, thereby demonstrating its effectiveness.
The code and dataset are available on our project site.
\end{abstract}

\vspace{-0.5cm}
\section{Introduction}
\begin{figure}
\begin{center}
    \begin{tabular}[b]{c}
      \includegraphics[height=2.9in]{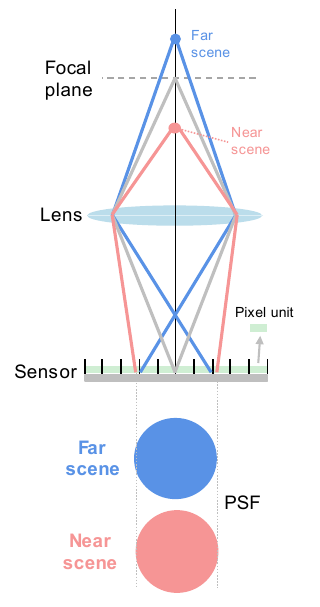}\\
      \small (a) Traditional sensor
    \end{tabular}
	\hspace{-0.8cm}
    \begin{tabular}[b]{c}
      \includegraphics[height=2.9in]{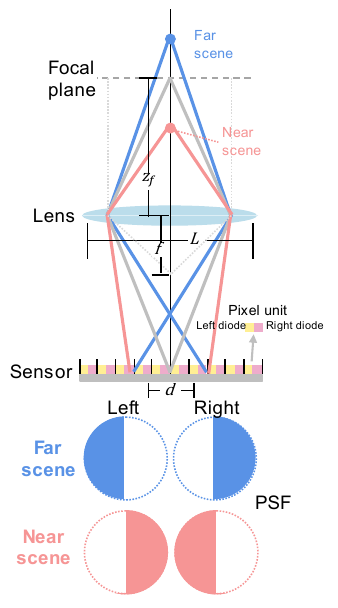}\\
      \small (b) Dual-pixel sensor
    \end{tabular}
\end{center}
\vspace{-0.5cm}
\caption{\textbf{PSF differences between traditional and DP sensors.}
(a) With traditional sensors, the shape of the PSF is the same in both far and near scenes, and the same defocus blur occurs.
(b) With the DP sensor, the PSF is shaded in the left and right halves of the left and right images, and its shape is inverted between the far and near scenes. This characteristic enables the DP sensor to calculate disparity by itself.
}
\label{fig:sensor}
\end{figure}

\vspace{-0.1cm}
\label{sec:intro}
Dual-pixel (DP) sensors~\cite{machado2013canon, fontaine2015state, okawa20191, yoon2021world, jung20221} have traditionally been used for camera autofocusing~\cite{herrmann2020learning, choi2023exploring}.
This sensor splits each pixel into two photodiodes, providing two views of the same scene, as shown in Fig.~\ref{fig:sensor} (b).
The DP view has a disparity that correlates with the amount of defocus blur. Therefore, moving the lens position to minimize the disparity allows the scene to come into focus quickly.
Similar to stereo cameras~\cite{barnard1982computational}, the disparity can also be converted into depth, and the ease with which a depth map can be acquired without active-light illumination has led to its use in applications such as depth-of-field extension~\cite{abuolaim2020defocus, xin2021defocus, abuolaim2022improving, yang2023k3dn} and segmentation~\cite{wadhwa2018synthetic}.
Because of their affinity for these applications, DP sensors are utilized in mobile cameras, such as Google Pixel, which require lightweight systems.

Many techniques have been proposed for DP sensor disparity estimation; however, most of them use end-to-end estimation with deep learning, as shown at the top left of Fig.~\ref{fig:teaser} (a).
Although this method is simple and direct, it does not fully exploit the disparity constraint, which is highly reliable for disparity near textures. This unnecessarily increases the network size and limits performance; specifically, it is prone to disparity artifacts correlated with RGB textures, as shown in Fig.~\ref{fig:result} (c).
As a na\"{i}ve solution, traditional template matching (TM)~\cite{brunelli2009template, lewis1995fast} can explicitly calculate disparity only in regions of high reliability. A lightweight network optimized for completion can complement the disparity, thereby minimizing the network size and suppressing artifacts.
However, the prominent anisotropic blurring property of DP outside the depth of field causes TM errors, which are further propagated by completion, resulting in degraded results.
There is also the problem of disparity errors near the edges owing to the combination of lightweight completion networks, and the expansion of disparity in principle owing to TM, which causes the side effects of intermediate value disparity even when simple filter-based refinement processes are applied.

Therefore, we propose a physics-informed lightweight disparity estimation method that explicitly exploits the disparity property of DP, as shown at the bottom of Fig.~\ref{fig:teaser} (a).
Specifically, we propose an approach that improves accuracy by modeling the TM errors arising from the physical and systemic properties of DP and applying appropriate sampling during training.
The physics DP simulator simulates varying optical conditions and fits parametric functions by using symbolic regression to efficiently model the TM errors.
Furthermore, we propose a framework that effectively compensates for disparity expansion errors that occur in principle~\cite{gupta2013window, hamzah2016literature} by carefully refining the network's output confidence map based on a filtering process.
The proposed non-learning-based disparity refinement framework is generally applicable and can be applied to the estimation results of other models.

In addition, a DP dataset is required to train the network; however, no dataset of DP images containing the actual disparity value exists to date~\cite{li2023learning}.
Therefore, conventional methods use depth as a reference and learn by devising a loss function to absorb the differences in the domain.
However, the network performance is limited by differences in the domain, and obtaining a large DP dataset with the ground truth of depth requires time and workforce, which are also limited in number and scale.
By converting the depth of widely available monocular RGBD datasets to DP disparity, we enabled large-scale training in the same domain without needing large amounts of DP data.

The performance of the proposed method exceeds that of the state-of-the-art method by less than $1/5$ of the parameters, demonstrating its superiority (Fig.~\ref{fig:teaser}(b)).
In summary, our contributions of this study are as follows:
\begin{itemize}
\item A practical method for estimating disparity (depth) from DP images in a lightweight and high-performance manner by modeling and appropriately constraining physics and systemic disparity properties. (Sec.~\ref{sec:problem},~\ref{sec:learning})
\item A non-learning-based refinement framework that efficiently handles disparity expansion errors by appropriately refining the confidence map that network output (Sec.~\ref{sec:refinement}).
\item An approach for learning disparity completion from common monocular RGBD data (Sec.~\ref{sec:learning}).
\end{itemize}

\section{Related works}
Wadhwa et al.~\cite{wadhwa2018synthetic} first attempted to estimate the disparity between DP images obtained using DP sensors. They used stereo matching to calculate the disparity and stabilized it with bilateral smoothing.
Punnappurath et al.~\cite{punnappurath2020modeling} proposed a DP-specific point spread function (PSF) to reproduce the defocus blur that occurs in DP sensors. They employed an optimization process to obtain the depth estimation.
However, these studies were based on simple filters, rule-based processing, and optimization,  limiting their disparity or depth estimation performance.
In recent years, end-to-end processing using deep learning has become mainstream, directly estimating disparity or depth from DP left and right images.
Garg et al.~\cite{garg2019learning} proposed a learning method that considers the indeterminacy inherent in DP. Pan et al.~\cite{Pan_2021_CVPR} proposed an unsupervised method that incorporated a physics DP simulator into the learning process.
Kim et al.~\cite{kim2023spatio} proposed a loss function that considers DP-specific symmetries and achieved state-of-the-art performance by tuning a pretrained stereo network.
Other studies have added new devices to DP sensors, such as coded-amplitude masks~\cite{ghanekar2024passive} and stereo cameras~\cite{zhang20202}, to improve the accuracy.
Although the deep-learning end-to-end processing used in these studies has shown impressive quantitative performance, it has limitations.
This approach tends to increase the network size due to its inability to leverage the highly reliable disparity constraint near textures.
Moreover, the internal processing of the network is a black box, which can lead to unexpected artifacts.
Therefore, we propose an efficient and effective estimation method that explicitly exploits the disparity constraint and leverages the physical properties of DP.
\begin{figure}
\includegraphics[width=\linewidth]{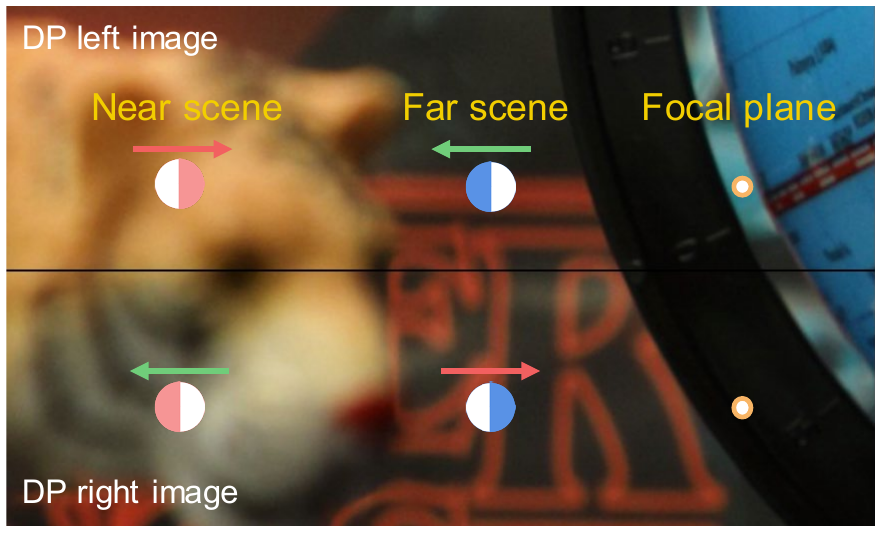}\\
\vspace{-0.5cm}
\caption{\textbf{Left and right images in a DP sensor.} The disparity generated by the DP sensor is small; however, the defocus blur is large. The PSF shape is different for left/right and perspective, making it more difficult to calculate disparity than when using a stereo camera.}
\label{fig:dp_images}
\vspace{-0.3cm}
\end{figure}

\section{Dual-pixel theory}
\subsection{Relation between disparity and depth}
\label{sec:relation}
To understand how the disparity and depth generated by the DP sensor are related, consider imaging a point-light source behind the focal plane and a point-light source in front of the focal plane in traditional and DP sensors, respectively, as shown in Fig.~\ref{fig:sensor}.
For a point light source located farther from the focal plane (far scene), the light passes through the primary lens and comes into focus in front of the sensor, resulting in an out-of-focus image.
For a point light source in front of the focal plane (near scene), the light passes through the primary lens and is focused behind the sensor, resulting in an out-of-focus image of the sensor.
Using the traditional sensor shown in Fig.~\ref{fig:sensor} (a), the PSF shape does not change for the far and near scenes and the same defocus blur is generated.
In contrast, for the DP sensor, as shown in Fig.~\ref{fig:sensor} (b), light passing through the left half of the primary lens aperture passes through the microlens at an angle that directs it toward the right half of the pixel and hits the right half of the pixel.
Light passing through the right half of the lens hits the left half of the pixel in the opposite direction.
Therefore, the PSF is shaded in the left and right halves, and its shape is inverted between the far and near scenes.
The two images produced by the segmented pixels produce a defocus blur proportional to the aperture diameter $L$. The amount of blur is proportional to the disparity between the two images, and the final disparity $d$ can be formulated as follows:
\begin{equation}
d=\alpha \frac{Lf}{1-f/z_\mathrm{f}}\bigg(\frac{1}{z_\mathrm{f}}-\frac{1}{z}\bigg),\\
\label{eqn:depth_to_disp}
\end{equation}
where $z$ is the depth of the object, $z_\mathrm{f}$ is the focal distance, $f$ is the focal length, and $\alpha$ is the proportionality constant.
$\alpha$ can be obtained by calibrating the chart in advance. If the camera parameters are known, the depth can be obtained from the disparity.

\subsection{Difficulty in calculating disparity in DP sensor}
\label{sec:difficulty}
Figure~\ref{fig:dp_images} shows examples of the left and right images acquired by the DP sensor, arranged vertically.
As shown in the figure, the disparity generated by the DP sensor is small because it is limited by the lens aperture $L$ in Eqn.~\ref{eqn:depth_to_disp}.
As the disparity increases, the defocus blur also increases, and the shape of the PSF differs depending on whether it is left or right and on the perspective.
Because of these properties, calculating disparity is more difficult for DP sensors than for stereo cameras.

\begin{figure}
\includegraphics[width=\linewidth]{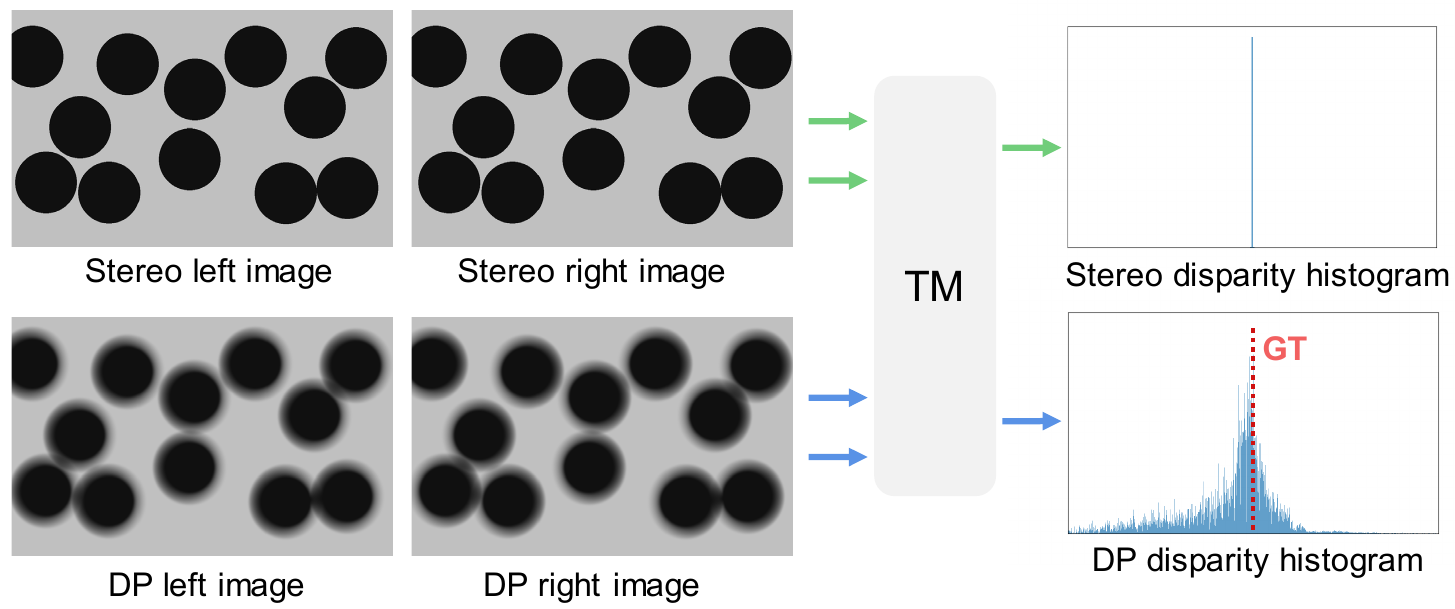}\\
\vspace{-0.4cm}
\caption{\textbf{Simulation of the difference between stereo and DP disparity calculation errors in the toy experiment.}
Histograms of disparity were obtained by TM for stereo and DP images using a random dot chart, respectively.
In stereo, the disparity is calculated accurately for most pixels; however, in DP, the disparity has many errors owing to the unique defocus blur.
}
\label{fig:toy}
\vspace{-0.3cm}
\end{figure}
\begin{figure*}[h]
\includegraphics[width=\linewidth]{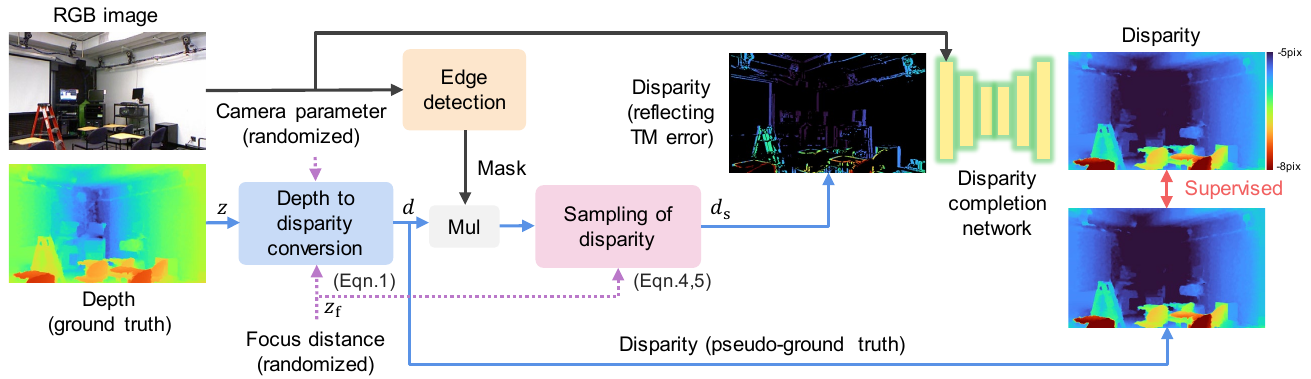}\\
\vspace{-0.5cm}
\caption{\textbf{Framework for learning disparity completion using RGBD data while incorporating the disparity properties of DP.}
After converting the ground truth from depth to disparity and extracting only the edge portions, input disparity reflecting DP properties is generated by sampling to achieve supervised learning.
}
\label{fig:training}
\vspace{-0.3cm}
\end{figure*}

To compare the difficulty of disparity calculation between stereo cameras and DP sensors, we performed a simple toy experiment, as shown in Fig.~\ref{fig:toy}.
We prepared random dot images from the stereo and DP cameras with the same ground-truth disparity.
DP images were generated using a physics-based DP simulator with arbitrary camera parameter settings (see supplementary material).
This simple example shows that TM can accurately calculate disparity for most pixels in standard stereo images.
However, in DP, the disparity error is significant because of the unique defocus blur.

\section{Method}
\subsection{Problem formulation}
\label{sec:problem}
In recent years, as shown at the top left of Fig.~\ref{fig:teaser} (a), the mainstream approach for disparity estimation in DP has been to learn a mapping $\mathcal{F}:\mathbf{I}_\mathrm{L}, \mathbf{I}_\mathrm{R} \rightarrow \mathbf{\hat{D}}$ such that disparity $\mathbf{\hat{D}}$ is directly estimated using the left and right images $\mathbf{I}_\mathrm{L}$ and $\mathbf{I}_\mathrm{R}$ of DP as inputs, as described in the following equation:
\begin{equation}
\mathbf{\hat{D}}=\mathcal{F}(\mathbf{I}_\mathrm{L}, \mathbf{I}_\mathrm{R}; \theta_\mathcal{F}),\\
\label{eqn:problem1}
\end{equation}
where $\theta_\mathcal{F}$ are the weights to be learned.

In principle, the disparity cannot be calculated in textureless regions and must be complemented by estimating the disparity from the surrounding textured regions.
However, end-to-end learning methods do not explicitly separate textured and nontextured regions, making the estimation process a black box. In many cases, the estimation results do not fully demonstrate the performance.
Additionally, performing estimation in a single network tends to make it redundant and bloated.

As a na\"{i}ve solution, assuming that the disparity is reliable in the region $\mathbf{M}$ near the edge where the texture is present, the disparity can be explicitly calculated in that region using template matching $\mathcal{T}$. Disparity can then be complemented by referring to the image.
The overview is shown at the bottom of Fig.~\ref{fig:teaser} (a).
Let $\mathcal{F}_\mathrm{c}:\mathbf{D} \rightarrow \mathbf{\hat{D}}$ be the mapping from the sparse disparity $\mathbf{D}$ to the dense disparity $\mathbf{\hat{D}}$, as follows:
\begin{equation}
\mathbf{\hat{D}}=\mathcal{F}_\mathrm{c}(\mathbf{D}, \mathbf{I}_\mathrm{L}; \theta_{\mathcal{F}_\mathrm{c}}), \quad \mathbf{D}=\mathbf{M} \odot \mathcal{T}(\mathbf{I}_\mathrm{L}, \mathbf{I}_\mathrm{R}).\\
\label{eqn:problem2}
\end{equation}
An edge-extraction filter can determine the confidence region $\mathbf{M}$.
By explicitly adding the disparity constraint, as in Eqn.~\ref{eqn:problem2}, the process can be divided into two parts: one calculates the disparity $\mathbf{D}$ of highly reliable texture regions, and the other uses reliable disparity $\mathbf{D}$ as an input to complete the disparity $\mathbf{\hat{D}}$ of texture-free regions by referring to the RGB image $\mathbf{I}_\mathrm{L}$.
The problem setup for disparity completion is similar to that for depth completion, allowing the use of sophisticated and efficient depth completion networks~\cite{nazir2022semattnet, yan2022rignet, wang2023lrru, tang2024bilateral, wang2024improving}, which have become increasingly competitive in recent years.
However, if the learned model for depth completion is applied directly to disparity completion, the performance is limited by domain gaps.
It would take considerable effort to obtain DP images containing the actual disparity value for training in the disparity domain, and no such dataset exists.
Furthermore, as discussed in Sec.~\ref{sec:difficulty}, errors in the disparity calculation of the DP sensor propagate throughout the completion process and thereby degrade the results.

\subsection{Disparity completion learning reflecting DP property using RGBD data}
\label{sec:learning}
\paragraph{Learning Framework Overview:}
We propose a physics-informed framework for learning disparity completion, as shown in Fig.~\ref{fig:training}, using widely available RGBD datasets such as NYUDv2~\cite{silberman2012indoor}, while considering the physics and system properties of DP.
First, the conversion in Eqn.~\ref{eqn:depth_to_disp} converts the ground truth from depth to disparity (which we define as pseudo-ground truth disparity), where the camera parameters are sampled randomly within a realistic range.
The edge extraction filter described in Sec.~\ref{sec:problem} is applied to the RGB image, and only the edge regions of the generated ground truth disparity are extracted to reflect the disparity properties of DP by sampling, considering the TM error in DP, which will be described later. In this case, the focus distance is the same as that in Eqn. ~\ref{eqn:depth_to_disp}.
The network can be trained in a supervised manner using the disparity of the edge regions, which are generated as input and supervised by the pseudo-ground truth disparity.

\begin{figure}
\includegraphics[width=\linewidth]{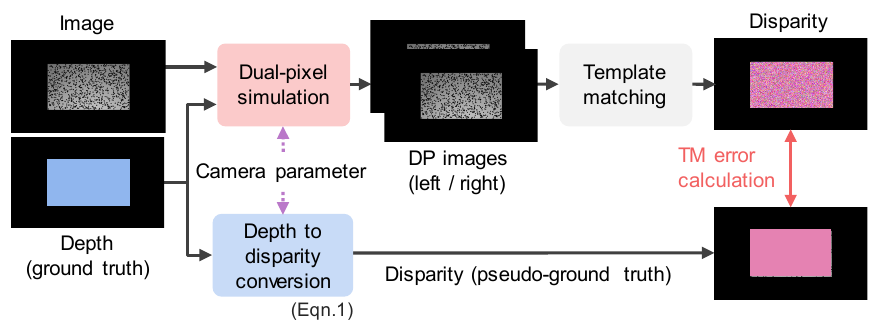}\\
\vspace{-0.2cm}
\caption{\textbf{Simulation for calculating TM errors in DP images.}
Input random dot chart and ground truth depth, and generate DP left and right images by a physics simulation reflecting the optical properties of DP.
Then, the disparity is calculated for the left and right images of DP by TM and compared with the pseudo-ground truth disparity to calculate the disparity error.
}
\label{fig:sim_disp_error}
\vspace{-0.1cm}
\end{figure}

\vspace{-0.2cm}
\paragraph{Error simulation for sampling reflecting DP property:}
We use a simulation to model the errors that occur in the TM of DP images.
As shown in Fig.~\ref{fig:sim_disp_error}, the left and right DP images are generated by a physics simulation~\cite{abuolaim2021learning} reflecting the optical properties of DP in which a random dot chart and ground truth depth are prepared in CG, specific camera parameters such as focus distance and f-number are set, and the PSF kernel is convolved according to the depth values.
The disparity is calculated for the generated left and right DP images by TM and then transformed from the depth of the ground truth to disparity using the same camera parameters from Eqn. ~\ref{eqn:depth_to_disp} to obtain the disparity of pseudo-ground truth.
The standard deviation of the disparity error is calculated by comparing the final calculated disparity with the pseudo-ground truth disparity.
\begin{figure}
\includegraphics[width=\linewidth]{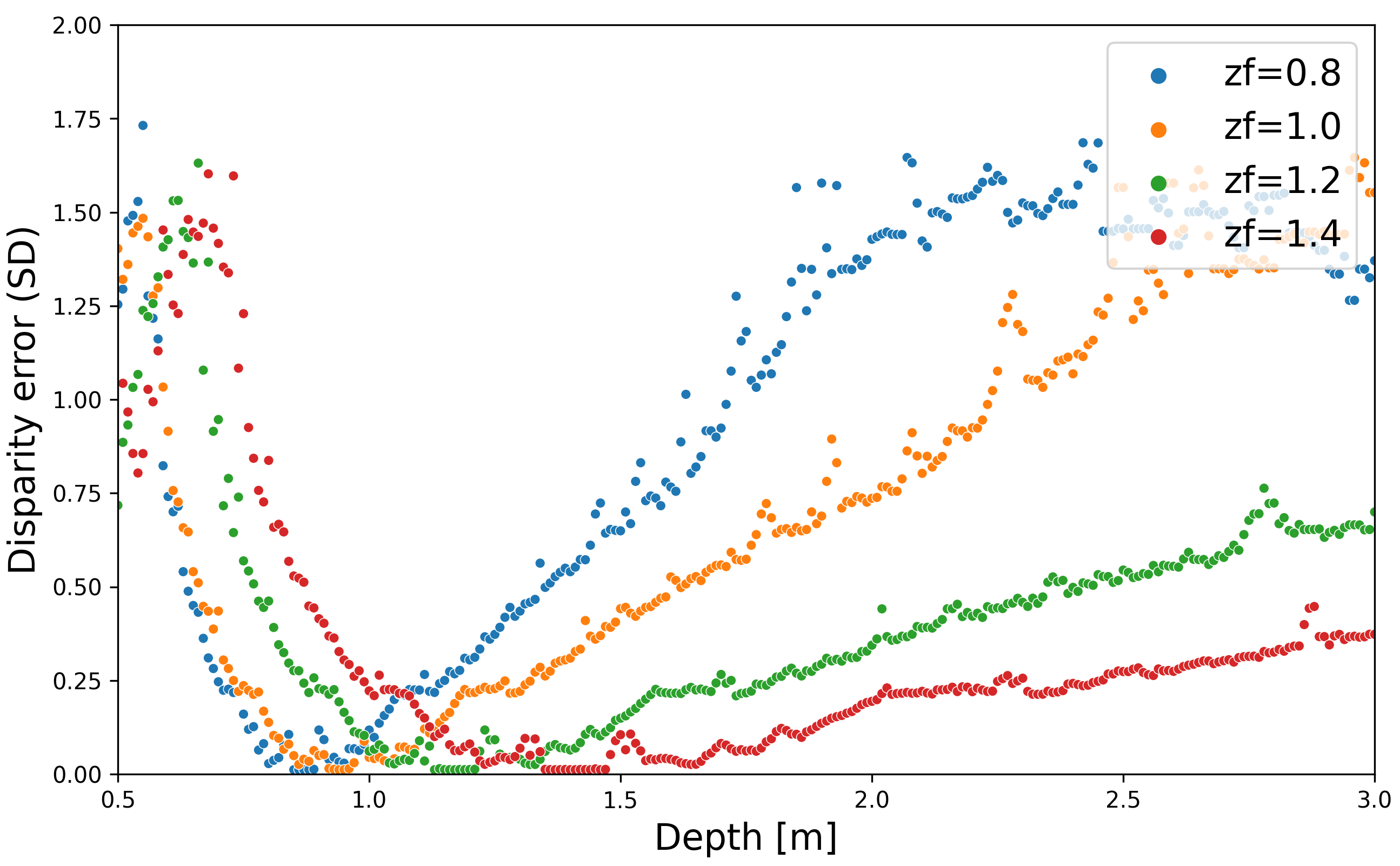}\\
\vspace{-0.6cm}
\caption{\textbf{Plots of TM error against depth for each focus distance.}
The horizontal axis is the depth of the ground truth ($z$), and the vertical axis is the standard deviation of the disparity ($\sigma_\mathrm{d}$), with colors indicating differences in focus distance ($z_\mathrm{f}$). When the focus distance equals the depth, the error approaches zero, whereas the error increases for front focus and rear focus.
}
\label{fig:disp_error}
\vspace{-0.1cm}
\end{figure}
This procedure is used to simulate and acquire data under various conditions by changing the depth and camera parameters.
Fig.~\ref{fig:disp_error} shows an example of the standard deviation of disparity at $f/2.0$.
The error approaches zero when the depth is equivalent to the focus distance and tends to increase when front and rear focuses are used.
We aim to model this standard deviation, which varies with the camera parameters, in an efficient and tractable form.
We used PhySO~\cite{tenachi2023deep}, a symbolic regression method that employs reinforcement learning for parametric modeling, to obtain a model equation for the standard deviation $\sigma_\mathrm{d}$ of the disparity concerning depth in the following equation:
\begin{equation}
\sigma_\mathrm{d} = c_{1}\Big(\frac{c_{2}z}{Fz_\mathrm{f}}\Big)^{\frac{z}{c_3}},
\label{eqn:disp_sd}
\end{equation}
where $c_1$,$c_2$, and $c_3$ are constants, $z$ is the depth of the ground truth, $z_\mathrm{f}$ is the focus distance, and $F$ is the F value.
This model equation is derived purely in a data-driven manner and has no specific physical basis.
We use the Laplace distribution~\cite{kotz2012laplace}, which is the best-fitting probability distribution for the distribution of errors obtained from the toy experiments shown in Fig.~\ref{fig:toy} to generate the disparity error for DP.
The final disparity is generated by adding an error term $n(\sigma_\mathrm{d})$ sampled from a zero-mean Laplace probability density function, as shown in the following equation:
\begin{equation}
d_\mathrm{s}=d+n(\sigma_\mathrm{d})
\label{eqn:probabilistic}
\end{equation}
where $d$ is the input disparity, and $d_\mathrm{s}$ is the sampled disparity.
This sampling can be used when training the disparity completion network to improve the completion accuracy because it reflects the TM error, which incorporates the DP's optical and system properties.

\subsection{Disparity refinement considering confidence}
\label{sec:refinement}
In principle, calculating the disparity using TM expands the disparity regions depending on the window size~\cite{gupta2013window, hamzah2016literature}.
In the case of DP images, because of the inherent defocus blur, a large window size (approximately 30 pixels) is required for matching, resulting in an excessive expansion of the disparity region, as shown in Fig.~\ref{fig:result_refinement} (c), Fig.~\ref{fig:result_nlspn} (b), and resulting in areas where the correct disparity cannot be calculated.
Therefore, despite perfect estimation by the completion network described in Sec.~\ref{sec:learning}, disparity errors will be significant near the edges.
In recent years, neural networks for refinement have been studied extensively; however, the problem is that the entire system becomes bloated because additional parameters are required.
For example, learning-based filtering methods, such as CSPN~\cite{cheng2018depth} and other refining methods~\cite{cheng2020cspn++, hu2021penet}, require more than 10 million parameters. In addition, these methods exhibit limited performance for error types not included in the training data, as observed in this case.
However, there are widely known general-purpose refinement methods, such as the joint bilateral filter (JBF)~\cite{kopf2007joint} and guided filter (GF)~\cite{he2012guided}, which do not require parameters.
However, correcting a wide range of errors with a realistic kernel size is difficult, and preventing intermediate disparities is challenging.

Therefore, we focused on the fast global smoother (FGS)~\cite{min2014fast}.
FGS can realize global optimization by combining 1D subsystems that reference neighboring pixels. It is suitable for efficiently refining a wide range of errors, such as colorization~\cite{levin2004colorization} and inpainting~\cite{bertalmio2000image}.
\begin{figure}
\includegraphics[width=\linewidth]{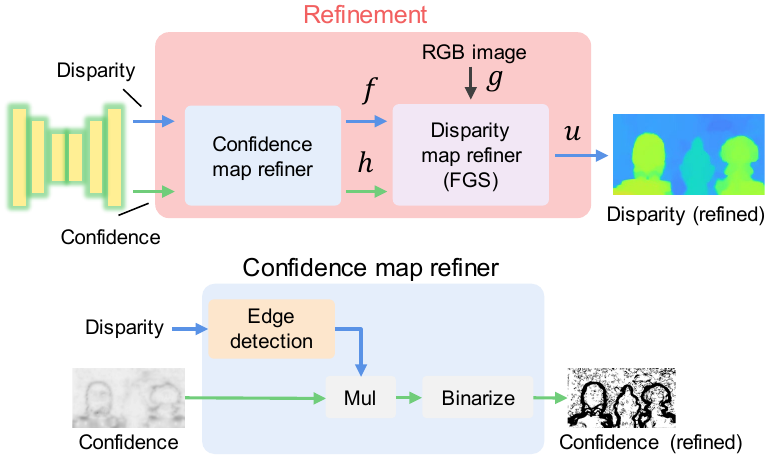}\\
\vspace{-0.4cm}
\caption{\textbf{
Refinement framework for correcting disparity expansion errors that occur in principle in TM process.
}
Pre-processing is applied to the disparity and reliability for input to the FGS, which also considers the reliability of the network output to reduce the impact of errors and artifacts, thereby improving stability.
}
\label{fig:refinement}
\vspace{-0.3cm}
\end{figure}
FGS is realized by solving the optimization problem expressed by the energy function in the following equation using a recursive filter:
\vspace{-0.1cm}
\begin{equation}
J(u)=\sum_p\bigg(h_p(u_p-f_p)^2+\lambda\sum_{q \in \mathcal{N}_4(p)}w_{p,q}(g)(u_p-u_q)^2\bigg),
\label{eqn:fgs}
\end{equation}
where $p$ is a pixel, $\mathcal{N}_4(p)$ represents the four neighbors of $p$, $u$ is the output data (refined disparity), $f$ is the input data (disparity to be refined), $h$ is the confidence map (binary) of the data term, and $w_{p,q}(g)$ is the similarity calculated from reference image $g$.

Figure~\ref{fig:refinement} illustrates the proposed refinement framework.
We propose improving the final disparity estimation performance by outputting the confidence map from the completion network and using it as a weight for the FGS.
Hence, we use the uncertainty-depth joint optimization loss function~\cite{zhu2022robust} to learn the uncertainty map unsupervised (the disparity is learned supervised).
The following equation expresses the loss function of the disparity completion network:
\begin{equation}
\begin{aligned}
&\mathcal{L}=\frac{1}{N}\sum \sqrt{e^{-s_i}(\hat{d}_i-d_{i}^{\mathrm{gt}})^2+2s_i},\\
&s_i=2\log \sigma_i, \quad e^{s_i}=\sigma_{i}^2,
\end{aligned}
\label{eqn:loss}
\end{equation}
where $d_{i}^{\mathrm{gt}}$ is the ground truth disparity, $\hat{d}_i$ is the complementary disparity, $N$ is the number of pixels, and $\sigma_i$ is the uncertainty map (inverted confidence map). As the uncertainty increases (lower confidence map), the agreement between the predictions and actual values is not required. Therefore, the learning process naturally leads to a lower confidence level in regions where variance is considered to be high.
By learning the confidence map simultaneously, the network focuses on learning more appropriate features, thereby improving completion performance.
The network configuration for outputting the confidence map is in the supplementary material.
We assume that regions near edges where disparity expands have low confidence and propose using edge detection for the disparity to refine the confidence map.
Specifically, the final confidence map is obtained by multiplying the edge detection results for the disparity output from the network by the initial confidence map and then binarizing to be input to the FGS.
To obtain the effect of FGS with fewer iterations, noise reduction pre-processing, a weighted median filter~\cite{ko1991center, ma2013constant}, is applied.

\section{Experiment}
\subsection{Dataset and implementation details}
We conducted several experiments to evaluate the proposed method.
For the evaluation, we used the 100-pair DP dataset from Punnappurath et al.~\cite{punnappurath2020modeling} and followed the same method used in previous studies to crop and evaluate the results.
TM was performed using a simple sum of absolute differences (SAD) process with a window size of 27 pixels and a search range of $\pm25$ pixels.
A mask for reliable disparity extraction was generated by applying a low-pass filter to the left image to suppress noise and then applying a process based on the Sobel filter~\cite{kanopoulos1988design}.
The datasets used to train the disparity completion network included  NYUDv2~\cite{silberman2012indoor} (1304 data points for training, 145 for validation), RGB-D-D~\cite{he2021towards} (4456 data points for training, 405 for validation), and our own CG data (9000 data points for training, 1000 for validation).
Three different datasets are mixed, shuffled, and trained at each epoch to prevent bias.
We used a network based on the cost volume-based depth completion network (CostDCNet)~\cite{kam2022costdcnet} with modifications, which is lightweight and performs well in terms of depth completion.
The network architecture was implemented using PyTorch~\cite{paszke2019pytorch} on a PC with an NVIDIA A100 GPU.
We trained the network for 50 epochs with a batch size of eight, Adam~\cite{kingma2014adam} as an optimizer, and an initial learning rate of 0.001. We then selected the model with the best validation results from the epoch.
The constants in Eqn.~\ref{eqn:disp_sd}, $c_1=6.93$, $c_2=0.48$, and $c_3=1.39$, were obtained by symbolic regression.

\subsection{Assessment}
As in existing studies~\cite{punnappurath2020modeling, Pan_2021_CVPR, kim2023spatio}, the affine invariant metric $\mathrm{AI}(p)$, calculated between the estimated disparity and the inverse depth of the ground truth, is used for evaluation.
$\mathrm{AI}(p)$ is defined as follows:
\begin{equation}
\mathrm{AI}(p)=\underset{\beta_0, \beta_1} {\operatorname{argmin}} \bigg( \frac{1}{N} \sum|d_{i}^{\mathrm{gt}}-(\beta_0+\beta_1 \hat{d}_i)|^p \bigg)^{1/p},
\label{eqn:ai}
\end{equation}
where $d_{i}^{\mathrm{gt}}$ is the inverse depth map of the ground truth, $\hat{d}_i$ is the estimated disparity, and $\beta_0$ and $\beta_1$ are the optimized affine coefficients.
The mean absolute error (MAE) for $p=1$ and the root mean squared error (RMSE) for $p=2$, with lower values indicating higher accuracy.
As in existing studies, we also evaluated $1-|\rho_\mathrm{s}|$ using Spearman's rank correlation coefficient between $d_{i}^{\mathrm{gt}}$ and $\hat{d}_i$ transformed using the optimized affine coefficients.
In this measure, the lower the value, the higher is the rank correlation between the two datasets.
Furthermore, the number of parameters in the network is simultaneously reported.

\paragraph{Comparison with conventional methods:}
\vspace{-0.2cm}
\begin{figure*}
\begin{center}
    \begin{tabular}[b]{c}
      \includegraphics[height=1.6in]{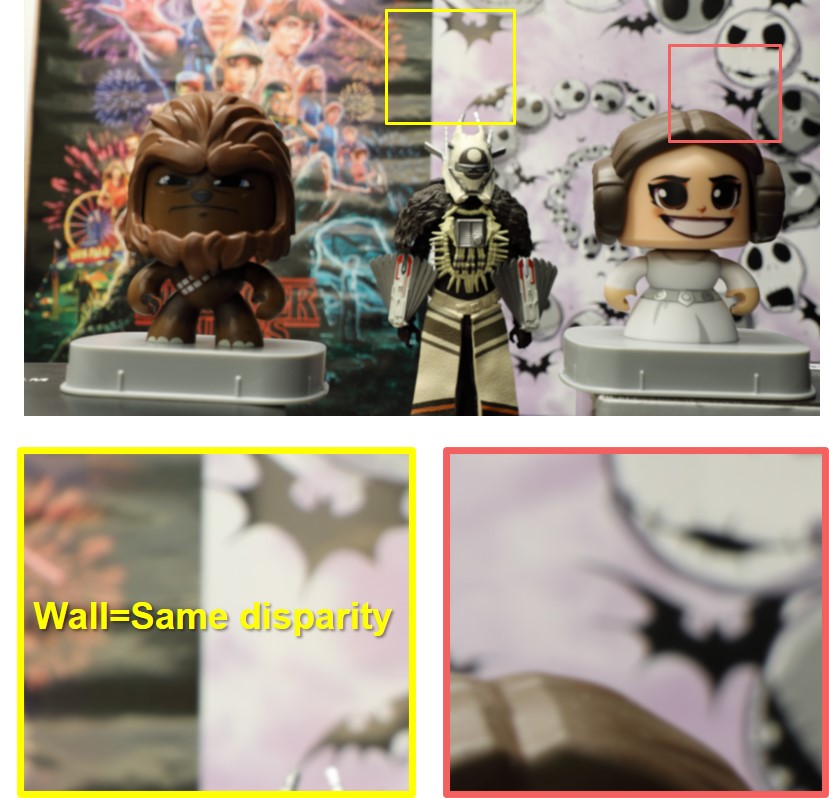}\\
      \small (a) RGB image
    \end{tabular}
	\hspace{-0.58cm}
    \begin{tabular}[b]{c}
      \includegraphics[height=1.6in]{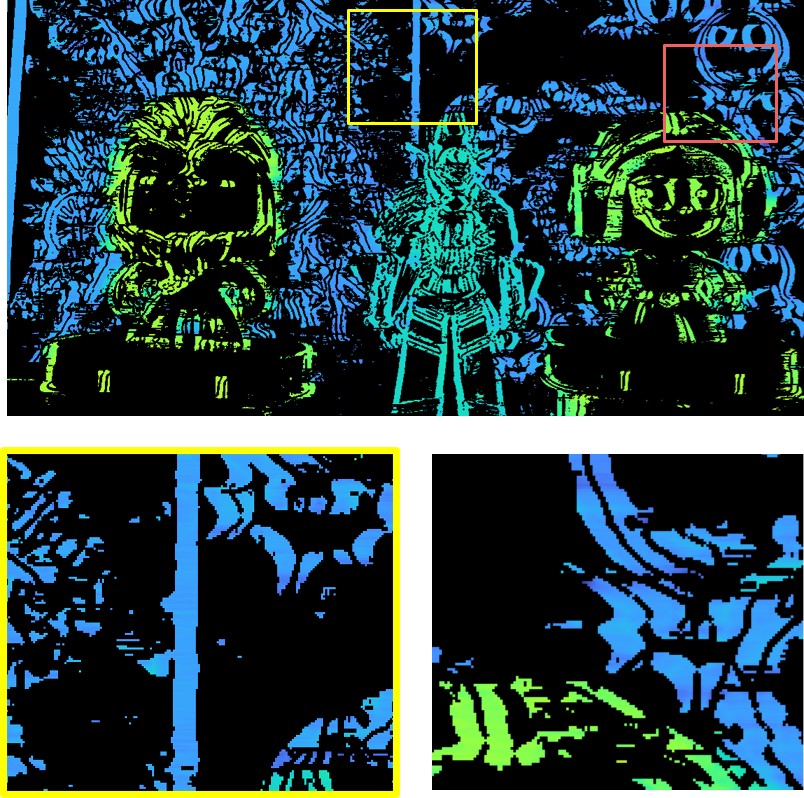}\\
      \small (b) Disparity map (TM)
    \end{tabular}
	\hspace{-0.58cm}
    \begin{tabular}[b]{c}
      \includegraphics[height=1.6in]{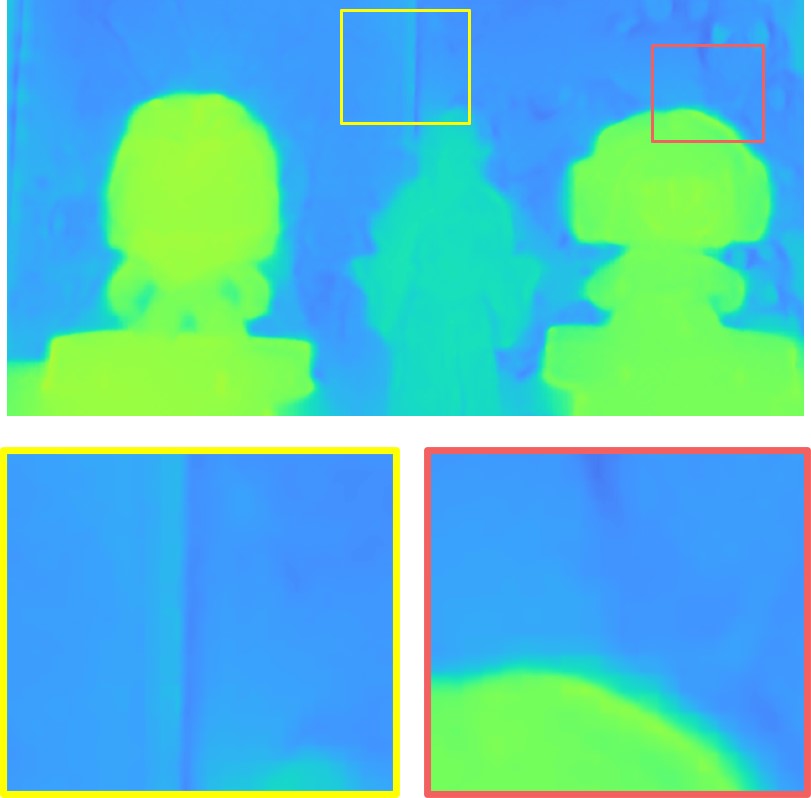}\\
      \small (c) Kim et al.~\cite{kim2023spatio}
    \end{tabular}
	\hspace{-0.58cm}
    \begin{tabular}[b]{c}
      \includegraphics[height=1.6in]{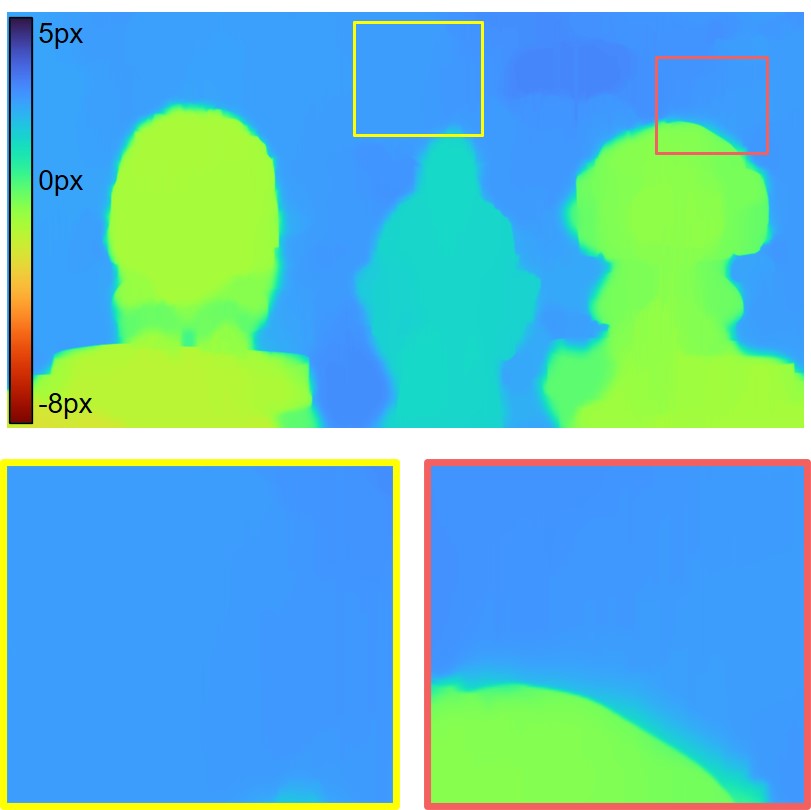}\\
      \small (d) Ours
    \end{tabular}
    \vspace{-0.2cm}
	\captionof{figure}{
	\textbf{Qualitative comparison of our results with other methods.}
 Kim et al.'s end-to-end method~\cite{kim2023spatio} produces false disparity in planes with uniform disparity (depth), such as walls. In contrast, our method does not generate such artifacts.
	}
\label{fig:result}
\end{center}
\vspace{-0.2cm}

\end{figure*}
\begin{table}[t]
  \caption{Disparity evaluation for each method on \cite{punnappurath2020modeling} dataset.}
  \vspace{-0.2cm}
  \centering
  \begin{tabular}{l|ccc|c}
    \toprule 
    Method & AI(1)$\downarrow$ & AI(2)$\downarrow$ & $1-|\rho_\mathrm{s}| \downarrow$ & Param \\
    \midrule
    Wadhwa~\cite{wadhwa2018synthetic} & 0.0463 & 0.0740 & 0.2875 & $\simeq0$\\
    Punnappurath~\cite{punnappurath2020modeling} & 0.0449 & 0.0724 & 0.2301 & $\simeq0$\\
    Pan~\cite{Pan_2021_CVPR} & 0.0894 & 0.1491 & 0.5008 & 11.0M\\
    Kim~\cite{kim2023spatio} & 0.0390 & 0.0679 & 0.2092 & 10.6M\\
    Ours & \textbf{0.0301} &\textbf{0.0667} & \textbf{0.0782} & 1.9M\\
    \bottomrule
  \end{tabular}
  \label{tbl:evaluation}
\end{table}
We compared our method with four other DP estimation methods: Wadhwa et al.~\cite{wadhwa2018synthetic}, Punnappurath et al.~\cite{punnappurath2020modeling}, Pan et al.~\cite{Pan_2021_CVPR}, and Kim et al.~\cite{kim2023spatio}.
Wadhwa et al.~\cite{wadhwa2018synthetic} used stereo matching followed by bilateral smoothing to obtain the disparity.
The other three methods directly obtain the disparity from DP images using an optimization process~\cite{punnappurath2020modeling} or a neural network~\cite{Pan_2021_CVPR, kim2023spatio}.
The results summarized in Tab.~\ref{tbl:evaluation} show that our method outperforms existing methods in the three evaluation metrics.
The number of parameters is also less than $1/5$ of other neural network-based methods, indicating that a lightweight and highly accurate estimation can be performed.
Figure~\ref{fig:teaser}(b) plots the number of parameters on the horizontal axis and performance on the vertical axis.
The closer to the lower left of the graph, the fewer the parameters and the better the performance, showing that our method achieves high performance while maintaining a low number of parameters.
Figure~\ref{fig:result} shows the results of the qualitative comparison.
The results show that Kim et al.'s end-to-end method generates false disparities in planes with uniform disparity (depth), such as walls. In contrast, our method does not produce such artifacts because it appropriately complements reliable disparities while considering DP properties.
In addition, our method generates sharper results even at the edges of the disparities without generating intermediate disparities.
\begin{table}[t]
  \caption{Ablation study.}
  \vspace{-0.2cm}
  \centering
  \begin{tabular}{l|cccc}
    \toprule 
    Measure & AI(1)$\downarrow$ & AI(2)$\downarrow$ & $1-|\rho_\mathrm{s}| \downarrow$\\
    \midrule
    Baseline (DC network)& 0.0477 & 0.0812 & 0.1436\\
    + (1) Sec~\ref{sec:learning} & 0.0476 & 0.0825 & 0.1427\\
    + (2) Sec~\ref{sec:learning} w/ Eqn.~\ref{eqn:disp_sd},~\ref{eqn:probabilistic}& 0.0381 & 0.0704 & 0.1121\\
    + (3) Sec~\ref{sec:refinement} & 0.0317 & 0.0684 & 0.0816\\
    + (4) Sec~\ref{sec:refinement} w/ conf refiner& \textbf{0.0301} &\textbf{0.0667} &\textbf{0.0782}\\
    \bottomrule
  \end{tabular}
  \label{tbl:ablation}
\end{table}
\begin{table}[t]
  \caption{Comparison of refinement methods.}
  \vspace{-0.2cm}
  \centering
  \begin{tabular}{l|c|ccc}
    \toprule 
    Method & Type & AI(1)$\downarrow$ & AI(2)$\downarrow$ & $1-|\rho_\mathrm{s}| \downarrow$\\
    \midrule
    w/o refinement & - & 0.0381 & 0.0704 & 0.1121\\
    JBF~\cite{kopf2007joint} & Filtering & 0.0365 & 0.0668 & 0.1121\\
    GF~\cite{he2012guided} & Filtering & 0.0355 &\textbf{0.0636} & 0.1005\\
    FBS~\cite{barron2016fast} & Filtering & 0.0378 & 0.0699 & 0.2302\\
    TGV~\cite{ferstl2013image} & Prior & 0.0368 & 0.0666 & 0.1041\\
    AR~\cite{yang2014color} & Prior & 0.0377 & 0.0697 & 0.2858\\
    CSPN~\cite{cheng2018depth} & Learning & 0.0380 & 0.0698 & 0.1130\\
    Ours & Filtering & \textbf{0.0301} & 0.0667 & \textbf{0.0782}\\
    \bottomrule
  \end{tabular}
  \label{tbl:disparity_refinement}
\end{table}

\paragraph{Ablation study:}
\vspace{-0.3cm}
Table~\ref{tbl:ablation} presents the results of an ablation study confirming the effectiveness of the proposed method.
The baseline directly uses a depth completion (DC) network trained on RGBD data to complete the disparity.
In (1), the conversion from depth to disparity described in Sec. ~\ref{sec:learning} using Eqn.~\ref{eqn:depth_to_disp} is performed during training.
In (2), in addition to (1), sampling is performed during training to reproduce the disparity error in template matching of DP images using Eqn.~\ref{eqn:disp_sd},~\ref{eqn:probabilistic} in Sec.~\ref{sec:learning}.
In (3), the refinement process of the disparity expansion error described in Sec.~\ref{sec:refinement} is added to (1) and (2).
In (4), the confidence map refinement and pre-processing are performed in addition to (1), (2), and (3).
The effectiveness of each measure was confirmed, and as can be seen from the results in (2), even without the refinement process of the disparity expansion error, this method performs as well as or better than the conventional methods in Tab.~\ref{tbl:evaluation}.
\begin{figure*}
\begin{center}
	\vspace{-0.2cm}
    \begin{tabular}[b]{c}
      \includegraphics[height=1.04in]{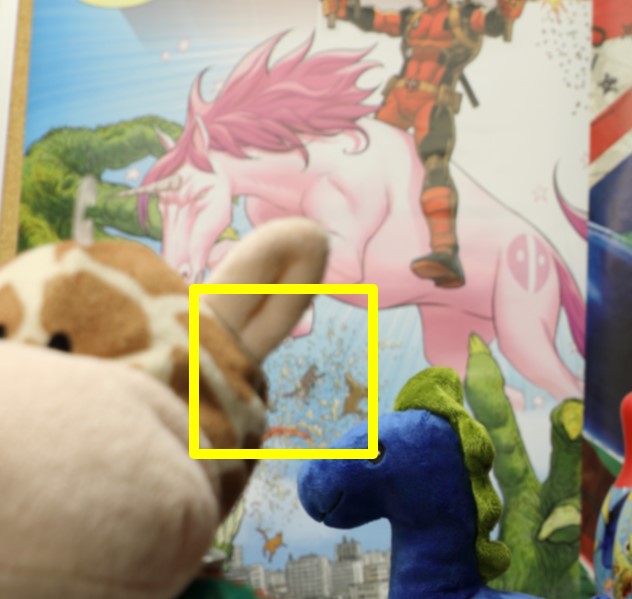}\\
      \small (a) Scene
    \end{tabular}
	\hspace{-0.58cm}
    \begin{tabular}[b]{c}
      \includegraphics[height=1.04in]{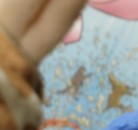}\\
      \small (b) RGB image
    \end{tabular}
	\hspace{-0.58cm}
    \begin{tabular}[b]{c}
      \includegraphics[height=1.04in]{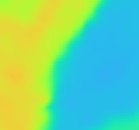}\\
      \small (c) w/o refinement
    \end{tabular}
	\hspace{-0.58cm}
    \begin{tabular}[b]{c}
      \includegraphics[height=1.04in]{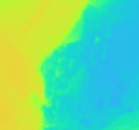}\\
      \small (d) GF~\cite{he2012guided}
    \end{tabular}
	\hspace{-0.58cm}
    \begin{tabular}[b]{c}
      \includegraphics[height=1.04in]{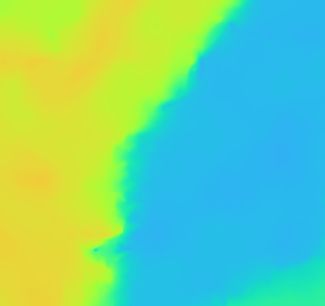}\\
      \small (e) CSPN~\cite{cheng2018depth}
    \end{tabular}
	\hspace{-0.58cm}
    \begin{tabular}[b]{c}
      \includegraphics[height=1.04in]{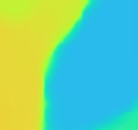}\\
      \small (f) Ours
    \end{tabular}
    \vspace{-0.2cm}
	\captionof{figure}{
	\textbf{Qualitative comparison of refinement methods.}
 GF produces intermediate disparity values correlated to the wall textures; however, our method does not produce such artifacts. Our method also has more precise boundary areas than the learning-based CSPN.
	}
\label{fig:result_refinement}
\end{center}
\vspace{-0.7cm}
\end{figure*}

\paragraph{Comparison with disparity expansion errors refinement:}
\vspace{-0.3cm}
\begin{figure}
\begin{center}
    \begin{tabular}[b]{c}
      \includegraphics[height=0.825in]{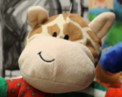}\\
      \small (a) RGB image
    \end{tabular}
	\hspace{-0.58cm}
    \begin{tabular}[b]{c}
      \includegraphics[height=0.825in]{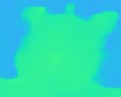}\\
      \small (b) w/o refinement
    \end{tabular}
	\hspace{-0.58cm}
    \begin{tabular}[b]{c}
      \includegraphics[height=0.825in]{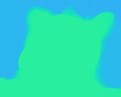}\\
      \small (c) Ours
    \end{tabular}
    \vspace{-0.4cm}
	\captionof{figure}{
	\textbf{Proposed refinement framework can be applied generically to other networks.} (b) Results of using NLSPN~\cite{park2020non}, a model with more parameters, as a completion model.
(c) Result of applying our proposed framework to (b).
	}
\label{fig:result_nlspn}
\end{center}
\vspace{-0.7cm}
\end{figure}
Table~\ref{tbl:disparity_refinement} shows how the effectiveness of our proposed framework for correcting disparity expansion errors compares to filter-based methods like JBF~\cite{kopf2007joint}, GF~\cite{he2012guided} and fast bilateral solver (FBS)~\cite{barron2016fast}, prior knowledge (model)- based optimization methods like the total generalized variation (TGV)~\cite{ferstl2013image} and adaptive autoregressive model (AR)~\cite{yang2014color}, and the learning-based method convolutional spatial propagation network (CSPN)~\cite{cheng2018depth}.
We integrated the CSPN after our completion network and trained it simultaneously within the framework shown in Fig.~\ref{fig:training}.
The results of these correction methods demonstrate the high accuracy of the proposed method.
For $\mathrm{AI}(2)$, GF was the best, however, the qualitative results in Fig.~\ref{fig:result_refinement} show that the GF results exhibit a significant intermediate disparity around the edges, whereas the proposed method exhibits no such side effects and is stable.
Learning-based methods, such as CSPN, exhibit limited performance in situations in which unexpected errors (disparity regions magnified by TM) are not included during training.
\paragraph{Apply our refinement framework to other disparity estimation models:}
\vspace{-0.3cm}
The proposed disparity refinement framework is sufficiently versatile to be applied to the disparity estimation results in other models.
Fig.~\ref{fig:result_nlspn} shows the results obtained by using a non-local spatial propagation network (NLSPN)~\cite{park2020non}, which is a model that has more parameters than does CostDCNet, as a completion model and applying our disparity refinement framework.
Even with a richer model, there is still some instability around the disparity edges. However, the refinement process can still estimate the disparity with high accuracy.
Please refer to the supplementary material for detailed comparative results, including the network configuration for outputting confidence in the NLSPN and for quantitative evaluation.

\section{Discussions and conclusion}
This study proposes a low-parameter and highly accurate method for disparity (depth) estimation from DP images.
Compared to conventional end-to-end processing, the proposed method is easy to implement in hardware because each processing phase and purpose are clearly separated.
The development of this field will improve the accuracy of the depth obtained from the DP sensor and expand its range of applications.
We will further study this method for future applications when an increasing number of cameras are equipped with a DP sensor.

\vspace{-0.4cm}
\paragraph{Limitations:}
In this study, to make a fair comparison with existing studies, an evaluation was performed on a specific dataset, and a detailed evaluation of data obtained in other settings (lens aperture, focal length, etc.) was not performed.
The proposed method assumes and assigns various optical parameters when considering DP properties, and we confirmed that it is qualitatively adequate on our own dataset (see supplementary material). A more exhaustive evaluation will be the subject of future studies.

{\small
\bibliographystyle{ieee_fullname}
\bibliography{egbib}
}

\end{document}